\documentclass{article}
\usepackage{ijcai23}

\usepackage{times}
\usepackage{soul}
\usepackage{url}
\usepackage[hidelinks]{hyperref}
\usepackage[utf8]{inputenc}
\usepackage[small]{caption}
\usepackage{graphicx}
\usepackage{amsmath}
\usepackage{amssymb}
\usepackage{amsthm}
\usepackage{booktabs}
\usepackage{algorithm}
\usepackage{algorithmic}
\usepackage[switch]{lineno}
\usepackage{float}
\usepackage{stfloats}
\usepackage{xcolor} 
\usepackage{bbding}
\usepackage{scalerel}
\usepackage{mathtools}

\DeclareMathOperator*{\argmax}{arg\,max}

\title{Pseudo-Labeling Enhanced by Privileged Information\\ and Its Application to In Situ Sequencing Images}

\author{
Marzieh Haghighi
\and
Mario C. Cruz\and
Erin Weisbart\and
Beth A. Cimini\and
Avtar Singh\and
Julia Bauman\and 
Maria E. Lozada\and
Sanam L. Kavari\and
James T. Neal\and Paul C. Blainey\and \\Anne E. Carpenter\And Shantanu Singh
\affiliations
Broad Institute of MIT and Harvard, USA\\
}

\begin{document}
\maketitle
\begin{abstract}
Various strategies for label-scarce object detection have been explored by the computer vision research community. These strategies mainly rely on assumptions that are specific to natural images and not directly applicable to the biological and biomedical vision domains. For example, most semi-supervised learning strategies rely on a small set of labeled data as a confident source of ground truth. In many biological vision applications, however, the ground truth is unknown and indirect information might be available in the form of noisy estimations or orthogonal evidence. In this work, we frame a crucial problem in spatial transcriptomics - \textit{decoding barcodes from In-Situ-Sequencing (ISS) images} - as a semi-supervised object detection (SSOD) problem. Our proposed framework incorporates additional available sources of information into a semi-supervised learning framework in the form of privileged information. The privileged information is incorporated into the teacher's pseudo-labeling in a teacher-student self-training iteration. Although the available privileged information could be data domain specific, we have introduced a general strategy of pseudo-labeling enhanced by privileged information (PLePI) and exemplified the concept using ISS images, 
as well on the COCO benchmark using extra evidence provided by CLIP.

\end{abstract}

\section{Introduction}

In many real world applications, the type of available information for data samples is not direct and the existing indirect evidence cannot be exploited using methods developed for datasets with direct ground truth. Creative exploitation of this side information into a learning framework might significantly advance real world applications, especially in life sciences. These applications often have unique properties that require unique strategies. Nevertheless, these domain specific strategies could provide potential generic insights applicable to other fields. This work is motivated by a problem called barcode calling, in which we attempt to decode information from In Situ Sequencing (ISS) microscopy images.

In situ sequencing methods are rising in popularity across genomics due to their application in spatial transcriptomics and optical pooled profiling screens. 
Their ability to preserve the spatial information of biological measurements while genotyping single cells at scale without the need to physically harvest and dissociate cells, adds rich information to profiles of biological samples~\cite{staahl2016visualization,vickovic2019high,asp2019spatiotemporal}. 
The field is exploding with new and improved genomic perturbation and spatial barcoding technologies. 
These technologies create a series of images to be analyzed computationally for recovery of the experiment's encoded barcodes. Each barcode is a sequence of base letters that are encoded in a sequence of images (each letter is an object in each image). Each barcode is associated with a cell in the experiment. 

Numerous attempts have been made to maximize the accurate recovery of barcodes over the past few years (see Section~\ref{subsec:rwiss}). 
Due to the simple structure of information encoded in these images and the lack of existent ground truth 
, traditionally barcode calling pipelines are formed by a series of expert-supervised multi-step image processing steps. However, they not only need expert optimization of parameters, but also are far from achieving maximal accurate detection of barcodes. 
Improvements would be extremely valuable, considering the cost and effort involved in performing extra experiments to compensate for missed signal and therefore statistical power of the downstream analysis. Recently, new methods \cite{chen2021barcode,andersson2021istdeco,gataric2021postcode} have attempted to model the physical process and find the optimum fit to an expected set of barcode solutions which is called an experiment's \textit{codebook}. 
A codebook tells us the set of possible barcodes existing (at unknown locations) in a set of images.
 It is thus an \textit{indirect} source of ground truth, which may be treated as side-information.
It is highly labor-intensive to generate even a small amount of \textit{direct} ground truth 
for this problem, because it would effectively require generating distinct datasets for each barcode, which is impractical except at a very small scale. Also, human observers cannot perform the task accurately on challenging instances, in light of the noise sources. 
The absence of such ground truth also limits strategies for evaluating performance and prevents consistent and fair comparison of new methodologies by well-defined metrics on benchmark datasets.

 In this work, we introduce a novel approach for framing the barcode calling problem, which also contributes to the deep semi-supervised learning and object detection literature. 
 The proposed framework takes advantage of special domain specific characteristics of this problem and dataset to form a novel learning strategy. 
We propose forming noisy labels by computationally cheap operations and then exploiting extra evidence available on data into the pseudo-labeling process in a semi-supervised object detection (SSOD) framework. This evidence, which could be in various forms and not limited to the examples we demonstrate here, are then used to adjust the initial decision boundary formed by noisy labels, during the self-training iterations. We name the overall strategy of incorporating privileged information in the pseudo-labeling process, \textit{Pseudo-Labeling Enhanced by Privileged Information}, and the algorithm specifically developed for barcode calling (that relies on the application of this strategy into a SSOD framework) is called PLePI-ISS.

We can summarize our contributions as:
\begin{itemize}
\item We propose PLePI as a novel strategy for incorporating complementary extra available information (termed `privileged information') for a problem into the pseudo-labeling process. This strategy is shown to not only salvage a model suffering from noise overfitting but also enhance the regular pseudo-labeling process in a semi-supervised learning framework.

\item We frame the crucial problem of barcode calling in the spatial transcriptomics field as a semi-supervised object detection problem with noisy labels and available privileged information. We also provide a public benchmarking resource which can be used for evaluation of the novel methodologies addressing this problem. The introduced framework, PLePI-ISS, which is the first end-to-end nonlinear framework to address the barcode calling problem, is evaluated in an out-of-sample fashion for the first time using the provided benchmark (whereas previous learning-based methods have only been evaluated within sample). 

\end{itemize}

%%%%%%%%% BODY TEXT
\section{Related Works}
\label{sec:related}
%-------------------------------------------------------------------------
\subsection{In Situ Transcriptomics Decoding Algorithms}
\label{subsec:rwiss}
Decoding barcodes from In-Situ Sequencing images 
 or \textit{barcode calling} is defined as reading sequences of cDNA copies of mRNA fragments in an image-based experiment.
 These fragments are spots scattered at various image locations. 
Each barcode is a length $N_r$ sequence of four A, T, C, G nucleotides, or `base letters', where $N_r$ is typically 9-16. 
A reference library of barcodes or \textit{codebook} (which is a subset of all possible combinations of length $N_r$ sequences of base letters) is designed and known before each experiment begins.

Each letter in the sequence is captured by one round or `cycle' of fluorescence microscopy images. 
Barcodes are formed by reading letters at a fixed location (or `spot'), in a series of images taken at consecutive sequencing cycles. 
The color of each letter (or `spot') corresponds to the specific intensity distribution of that spot across four base letter color channels. 
An example of a cycle image is shown in Supplementary Figure 1\footnote{\label{fn:sup} The source code and supplementary materials for this paper can be found at \url{http://broad.io/PLePIISS}}, in which the top row shows a zoomed single spot to be decoded to one of the base letters, based on its intensity in the four color channels corresponding to A, T, G, or C. 
Despite the structural simplicity of these ISS images, many practical complications exist, such as background from the cellular matrix, variable focus quality in thicker sample mounts, imbalances of brightness across color channels, bleed-through of signal from one color channel to another or from one cycle of imaging to another.  

Barcode calling algorithms are typically a multi-step expert-supervised process. 
Images are first corrected for general uneven microscopy patterns (due to uneven illumination or detection) and then aligned across channels and across rounds or cycles of imaging. 
The crucial step then is to recover the signal from noisy measurements. Signal here refers to the location of spots and their label. 
Previous literature has used various techniques to overcome this challenge by modeling and correcting the different sources of noise ~\cite{andersson2021istdeco,gataric2021postcode,chen2021barcode,senel2022optocoder}. 
Various bottom-up or top-down processing approaches have been proposed and implemented as pipelines in the open-source Starfish~\cite{axelrod2021starfish} Python library. 
Bottom-up or spot-based approaches first determine the existence of a spot and then estimate the barcode in the spot across all rounds of imaging \cite{shah2016situ,wang2018three,gyllborg2020hybridization}. 
On the other hand, top-down or pixel-based approaches look for the existence of a barcode in the experiment’s codebook to determine the signal versus noise ~\cite{andersson2021istdeco,chen2021barcode,lee2014highly,moffitt2018molecular}.
Among recent pixel-based decoding approaches, BarDensr models the generative physical process linearly, considering various factors including the probe response function, phasing effect, cross-talk, background noise, and the codebook of barcodes. 
It estimates model parameters from data and solves a sparse non-negative regression problem by constrained optimization~\cite{chen2021barcode}. 
ISTDECO suggests optimization of a single model for simultaneous barcode localization and decoding for further denoising and barcode calling improvement~\cite{andersson2021istdeco}. 
A more recent spot-based model called PoSTcode~\cite{gataric2021postcode} suggests probabilistic modeling of data by a matrix-variate Gaussian Mixture Model and estimates posterior probabilities of barcodes given the codebook and image measurements at spot locations. Without suitable ground truth, the field has not been able to compare methods against a unified benchmark. 

\subsection{Object Detection With Limited Annotations}

Deep learning-based object detection in the presence of ground truth annotations has shown outstanding success. However, the labor-intensive and costly nature of collecting annotations in object detection, has motivated the recent attempts to study various learning strategies with limited supervision. 
Weakly supervised object detection (WSOD) studies the strategies to use image-level categorical annotations without location information to detect objects~\cite{tang2018pcl,dong2021boosting,lin2020object,shen2020uwsod}. 
Semi-supervised object detection (SSOD) leverages a small set of fully annotated images together with a large set of unannotated images for detecting objects in the unannotated data. 
SSOD attempts have mainly followed variations of the well-established strategies in semi-supervised image classification such as pseudo-labeling and consistency regularization~\cite{jeong2019consistency,jeong2021interpolation,sohn2020simple,kuo2020featmatch}. 
However, due to the difficult nature of object detection relative to the image classification problem, the literature on SSOD is limited, and it lags behind semi-supervised object classification. 
Various other strategies have been also explored for label-scarce object detection; sparsely annotated object detection deals with partial annotations in each image \cite{wang2021co,niitani2019sampling,zhang2020solving}, single instance object detection takes annotations of a single instance per category ~\cite{li2022siod}, point-supervised object detection replaces box-supervision with single quasi-center point supervision for saving annotation costs~\cite{chen2022point}. These variations all aim to achieve a performance close to fully supervised object detection by using minimal/sparse levels of supervision. Our proposed strategy could benefit any categories of the work mentioned that involve pseudo-labeling where we have access to a source of privileged information.

%%%%%%%%% Methods
\section{PLePI}
\label{sec:PLePI}

\begin{figure*}[t]
  \centering
   \includegraphics[width=0.85\linewidth]{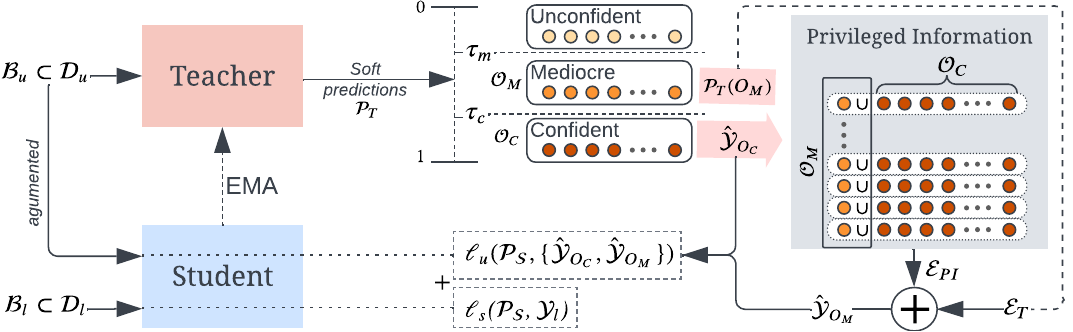}

   \caption{Schematic representation of PLePI in a teacher-student learning loop. The teacher takes each single sample’s soft predictions in a mini-batch of samples and assigns pseudo-labels to each sample using the available privileged information.}
   \label{fig:overview}
\end{figure*}

Here, we describe our proposed domain-free strategy for enhancing pseudo-labeling by a source of privileged information. Following the related works ~\cite{zhou2021instant,xu2021end,li2022pseco,zhou2022dense}, we build on a teacher-student scheme, where both teacher and student share identical architecture, and the teacher's weights are updated iteratively by exponential moving average (EMA) of the student's weights. The teacher model makes decisions based on weakly augmented unlabeled data and generates labels for strongly augmented data inputted into the student model. This process, which is called input consistency regularization, ensures gradual improvement of the student and therefore the teacher in a dual iterative learning scheme. Without loss of generality, we describe the enhanced pseudo-labeling algorithm for a single learning task and loss function. 

Assuming sets of labeled ($\mathcal{D}_l$) and unlabeled ($\mathcal{D}_u$) samples, the teacher model is first trained by the set of labeled data. The student model is trained by using both labeled and unlabeled sets to minimize a linear combination of, respectively, a supervised and an unsupervised loss; $\ell_s+\lambda_u \ell_u$, where $\lambda_u$ defines the contribution of unlabeled samples to the student's overall loss function. 
In each training iteration, the teacher model generates a set of class conditional probabilities $\mathcal{P}_{T}$ for each sample in a mini-batch of samples. In a typical pseudo-labeling by confidence thresholding approach, we pseudo-label 
samples where their likelihood of belonging to any class $k$ is higher than $\tau_c$. These pseudo-labels are then fed to the unsupervised loss to update the student model using a cross-entropy or focal loss which minimizes the distance between the student's soft predictions and the teacher's generated pseudo-labels. We call the samples selected for pseudo-labeling, the set of confident samples 
% ${O}_C = \{o |\mathcal{P}_{T}(y_o = k | \forall k \in K )> \tau_c \}$, 
${O}_C = \{o | \underset{{\scaleto{k\in K}{5pt}}}\max~ 
 {\mathcal{P}_{T}(y_o = k)}>\tau_c\}$, 
where $K$ refers to the full set of class labels. 
Here we define an additional \textit{mediocre} set of samples with lower levels of confidence ($\tau_m\le\tau_c$), ${O}_M = \{o |\tau_m\ge\underset{{\scaleto{k\in K}{5pt}}}\max~ 
 {\mathcal{P}_{T}(y_o = k)}\ge\tau_c \}$.
When privileged information is available, the teacher fuses its own soft predictions ($\mathcal{P}_{T}$) into each mediocre sample $o_m$ with the evidence the privileged information provides ($\mathcal{E}_{PI}$) on the label of that sample.
We model the probability of a sample label using both types of evidence as:
\begin{align}
\mathcal{P}(y|{\mathcal{E}_{T},\mathcal{E}_{PI})} \propto 
\mathcal{P}(y|{\mathcal{E}_{T})\mathcal{P}(y|\mathcal{E}_{PI})} 
\end{align}

The choice of $\tau_m$, $\tau_c$, and their relative distance could be dependent on the problem and context. For example, when we have a poorly calibrated model due to the overfitting to noisy labels, we can select $\tau_c=1$, which means that we fuse the privileged information into all samples passing a threshold according to the Teacher's soft decisions as teacher's confidence is not reliable. On the other hand, when we have a reliable teacher, we assume the teacher's confident decisions may not benefit from privileged information and $\tau_c$ could be set to lower values.

We assume privileged information could provide evidence or could be converted to the form of probabilities for a group of objects co-occurring in a mini-batch of samples. Hence, for a set of $\mathcal{O}=\{o_j\}_j$ of objects in a mini-batch $\mathcal{B}$, privileged information could provide $\{\mathcal{P}_{\scaleto{PI}{4pt}}(\mathcal{O}_S)| \mathcal{O}_S\subset\mathcal{O} \}$, where $\mathcal{O}_S$ could be any possible subset of objects in $\mathcal{B}$.

Then, we fuse the evidence provided by the current teacher's model ($\mathcal{E}_{T}$) and privileged information ($\mathcal{E}_{PI}$), to form decisions for each object in the mediocre set, as also shown in the schematic overview of the strategy in Figure~\ref{fig:overview}. Thus, for every sample in the mediocre set, $o_m \in \mathcal{O}_{M}$, we estimate its label $k^*$ as:
\begin{align}
k^* & =\underset{{\scaleto{k\in K}{5pt}}}\argmax~ 
 {\mathcal{P}(y_{o_m}=k | {\mathcal{E}_{T},\mathcal{E}_{PI})}} \nonumber \\
 &=\underset{{\scaleto{k\in K}{5pt}}}\argmax~ 
 {\mathcal{P}_{T}(y_{o_m}=k)
 \mathcal{P}(y_{o_m}=k|\mathcal{E}_{PI})
 }.
\nonumber \\
  &=\underset{{\scaleto{k\in K}{5pt}}}\argmax~ 
 {\mathcal{P}_{T}(y_{o_m}=k)
 \mathcal{P}_{\scaleto{PI}{4pt}}(o_m\cup\mathcal{O}_{C}|y_{o_m}=k)
 }.
\label{eq:k}
\end{align}

As mentioned, $K$ refers to the full set of class labels, however, we recommend limiting $K$ to a set of top $n$ candidate labels for each $o_m$ according to $P_T$ to avoid unnecessary computational overload. The algorithm in compact form is presented in Supplementary B\textsuperscript{\ref{fn:sup}}.

\section{PLePI-ISS}
\subsection{ISS Data-Structure}
\label{subsec:ISSds}
The images captured in each sample well of an experiment consist of $N_f$ field of views, $\{\mathcal{I}_f\}^{N_f}_{f=1}$, where each $\mathcal{I}_f$ is a sequence of images captured at $N_r$ rounds or cycles of the experiment; $\mathcal{I}_f=\{I_{f,r}\}^{N_r}_{r=1}$. Therefore, for each site of the experiment, there exist $N_f\times N_r$ number of $I \in\mathbb R^{W\times H\times C}$ images of height $H$, width $W$ with $C$ color channels. A barcode is formed by a sequence of $N_r$ objects at a specific fixed location across $\mathcal{I}_f$ images.

We frame the barcode calling problem as an object detection problem on single-cycle images in the presence of noisy labels and privileged information. 

\subsection{Domain Specific Properties}
Here we describe data properties that make our SSOD problem distinct and therefore provide the rationale for our design choices, which may not align with conventional SSOD techniques utilized for natural images.

\subsubsection{Unique Structure}
Our data has an unusual structure which is different from natural images that are the target of most SSOD designs. Compared to natural images, the objects' structure is much simpler which creates limitations on choices of augmentation and therefore regularization capacity which is required in a self-training learning strategy. Objects are very small spots in various forms densely scattered throughout each input image (Supplementary Figure 1\textsuperscript{\ref{fn:sup}}). The label of each spot is shape independent and manifested by the arrangement of intensity distribution across color channels. Slight perturbations in these arrangements could create noise in predictions. On the other hand, this simple structure might allow extracting noisy labels by cheap and efficient image processing techniques. The size of objects (4-10 pixels in each dimension) allows using point-level annotation which is cheaper to extract compared to box-level annotation.
Another unique property is the consistent location of objects across various cycles of the experiment (i.e. all objects corresponding to each barcode have similar location labels), which allows model regularization for the object detection task in lieu of intensity-based data augmentations.

\subsubsection{Self-training and Noisy Labels}
\label{subsec:selftrain}
The basic idea behind self-training is that a model can improve itself by learning from its own confident predictions~\cite{mclachlan1975iterative,scudder1965probability}. 
The confident predictions are then added as labeled samples to the next training iteration. The initial model that is called a “pseudo-labeler” is typically trained using a source of confident labels.
If a pseudo-labeler is fitted to noisy labels, the model would overfit to those labels and can’t rely on its own confident predictions to improve its learned decision boundary~\cite{arazo2020pseudo}. 
Strategies for alleviating noisy pseudo-labeling due to the confirmation bias in a poorly calibrated model are very limited~\cite{arazo2020pseudo,li2020dividemix,rizve2021defense} and domain-specific. The fusion of privileged information to the model's decision could be a remedy to this problem in our application.
By PLePI, a model learned from noisy labels corrects its formed decision boundary by adjusting pseudo-labels in a mini-batch of data using a source of privileged information. 
We emphasize that, although the proposed strategy can improve the quality of the pseudo-labels generated in any self-training loop, it is particularly beneficial to a model suffering from noise overfitting which otherwise fails to benefit from unlabeled data through self-training iterations.

\subsubsection{Codebook} As described in earlier sections, the experiment-specific codebook gives prior information on the expected set of solutions for the barcode calling problem. However, it doesn't provide any direct information on the label of each spot to be labeled by an object detector.

\subsection{Privileged Information in ISS Barcode Calling}
\label{subsec:PIISS}
According to $\{\mathcal{P}_{\scaleto{PI}{4pt}}(\mathcal{O}_S)| \mathcal{O}_S\subset\mathcal{O} \}$,
evidence provided by privileged information is increased when we have more sample subsets that can provide information about each other in the form of a group (within a mini-batch of samples). 
Therefore, to maximize the amount of information the teacher has access to in each training iteration, we form a mini-batch by grouping images that can provide the maximum information about each other. In ISS domain application, this translates to tiles of images across various cycles of a single field of view.

\paragraph{Shared evidence on object locations across images.} 
As mentioned earlier, the location of objects across images of various cycles in an experiment remains constant. Slight variations in object locations in these images are due to imaging noise and misalignments across cycles. 

We can exploit this data property as an evidence on the location of a group of spots, which states that the probability for a group of spots at a fixed $(x,y)$ location in various cycle's images is one if and only if the objects in the group have either all foreground or all background labels.

\paragraph{Codebook as an external source of evidence.} Labels for each foreground object at a specific location across all cycles form a barcode sequence which should exist in the reference barcode library of barcodes (the experiment’s codebook) in a noiseless scenario. 
However, due to experimental imaging noise, the formed barcode may not match any of the barcodes in the codebook, but it could still be assumed to remain in the neighborhood of its ground truth barcode in a Hamming space where all possible barcodes reside. 
Therefore, for an ordered set of object labels across cycles that form a barcode $B$, this evidence states $\mathcal{P}(B)=1$ only if $B \in codebook$ and otherwise $\mathcal{P}(B)=0$.

\subsection{PLePI-ISS: Application Into a Two-Stage SSOD}
In the previous section, we proposed arranging images in each mini-batch so that object detection solutions across images become dependent and form a source of privileged information. 
We described our problem’s domain-specific properties for the localization and classification tasks which are used as the sources of privileged information. 

Given $N_f\times N_r$ unlabeled images defined in~\ref{subsec:ISSds} across all the fields of view and cycles in a well of an ISS experiment, noisy estimates for location and categories of objects are extracted for a subset of images by: 1. simple automated thresholding based approaches (cheap but low quality) and 2. a multi-step expert-supervised image processing pipeline (still noisy but higher quality). 
We later evaluate the effect of annotation noise level on our proposed model's performance.

We apply our learning strategy into a
teacher-student based SSOD, build on the two-stage anchor-based Faster R-CNN~\cite{ren2015faster} object detector. Two-stage object detectors have specifically shown promising results in semi-supervised object detection ~\cite{zhou2021instant,wang2021data}.
A teacher model is then initialized using noisy labels. The teacher attempts to correct the “overfitting to noisy labels effect” by updating its decisions according to the privileged information.

The multi-task loss function for the Faster R-CNN~\cite{ren2015faster} two-stage object detector is a linear combination of object localization and classification losses. 
In the first stage, the region proposal network (RPN), is trained for binary classification of foreground versus background and bounding box regression of a set of predefined anchors. A region of interest (ROI) detection network is simultaneously trained for object classification and bounding box regression of a selected set of ROIs.
% For classification sparse cross-entropy is used and for bounding boxes , Smooth L1 Loss is used.
In each training iteration, the RPN generates a set of objectness scores and bounding box deltas for each proposal anchor in the image. The objectness scores are used to assign an anchor to foreground versus background class and bounding box deltas are to estimate the adjustment needed to the anchor bounding box for each detected object. 

Following the shared evidence on object locations across images (as described in~\ref{subsec:PIISS}), feature maps for extracting objectness score across different cycles of imaging in each tile of a single field of view that are arranged in a mini-batch, $\mathcal{I}_f=\{I_{f,r}\}^{N_r}_{r=1}$, can be considered as noisy perturbations of the true unknown feature map of locations. We minimize the inconsistency across these perturbed images, by enforcing all the objectness predicted labels (foreground versus background) for each anchor to be identical. The pseudo-labeling of each box on the feature map is done by the consensus-predicted labels across different perturbations. A confidence threshold for pseudo-labeling of RPN boxes to the foreground class is fixed and applied to consensus (median) objectness scores. Then, the median of estimated bounding box deltas for detected-as-foreground objects are calculated and considered as ground truth annotations for minimization of their corresponding cross-entropy and $L1$ loss functions of the RPN loss, $\mathcal{L}_{RPN}$ (Supplementary figure 3 (c)\textsuperscript{\ref{fn:sup}}).

Next, selected ROIs detected as foreground objects go to the second stage or the ROI detection network. We apply random rotations to the ROI feature maps in the student network. Each foreground anchor across all cycles' images forms a set of $N_r$ objects, $B$, with the probability of a sequence formed by their independent object labels being $\prod\limits_{i=1}^{N_r}\underset{{\scaleto{k\in \{A,T,C,G\}}{5pt}}}\max~ 
 {\mathcal{P}_{T}(y_{o_i} = k)}$. 
Next, the set of $N_r$ objects in each barcode is divided into two sets of confident and mediocre objects. Confident objects whose maximum class conditional probability is more than $\tau_c$ are pseudo-labeled and form the confident set of pseudo-labels $\hat{\mathcal{Y}}_{O_C}$.

Incorporation of the codebook as an external source of evidence (described in~\ref{subsec:PIISS}) into equation~\ref{eq:k} will result in pseudo-labeling of mediocre objects so that the final barcode formed by the group is the most probable barcode that also exists in the codebook. The cross entropy between the teacher's estimated and updated pseudo-labels $\{\hat{\mathcal{Y}}_{O_{C}},\hat{\mathcal{Y}}_{O_{M}}\}$ and the student's soft predictions $\mathcal{P}_{S}$ is then minimized by $\mathcal{L}_{ROI}$ to update the ROI detection network (Supplementary figure 3 (d)\textsuperscript{\ref{fn:sup}}).

\subsection{Barcode Calling Benchmarking}
\label{subsec:bcbench}
\subsubsection{In-Situ-Sequencing Benchmark Resource}
We created a benchmark resource for barcode calling by conducting an experiment testing a library of barcoded genetic reagents, which includes a plate of in-situ sequencing (ISS) images along with the corresponding next-generation sequencing (NGS) data as indirect ground truth. The library contains 186 barcodes of nine letters represented in nine cycles of an imaging sequence. The resource (described in detail at Supplementary A\textsuperscript{\ref{fn:sup}}) contains two six-well plates, each well containing about 100 images (called sites or fields of view). Each well contains about 500000 single cells and 20 million four-channel spots in ISS images of nine cycles. Not all cells have spots, and most cells have more than one spot. The results of per-spot barcode assignments based on the proposed and baseline methods are then converted to per-cell barcode assignments as described in the next section. The abundance of cell-level barcode assignments should approximate NGS-based barcode abundances. We use the latter, as the main reference to assess the quality of the former extracted by various decoding strategies.

\subsubsection{Cell Calling as a Proxy for Barcode Calling Evaluation}
\label{sec:cellcall}
Cell calling is defined as assigning a barcode to each cell, indicating the perturbation that the cell has received. There are usually multiple individual barcode spots detected in each cell. They ought to report identical sequences within a given cell (except when the experiment is configured otherwise), but that is not always the case due to confounding noise. Our experiments are configured to typically have a single perturbation per cell and therefore any inconsistency in the detected barcodes is likely due to a computational decoding error. For each cell, we need to make a decision on the cell’s perturbation (cell calling) based on the barcode calling evidence at hand. To map all methods’ barcode calling results to cell-level barcode assignments, we use cell-segmentation image outputs from a classical image processing pipeline using CellProfiler, which uses cytoskeleton-stained cell images to identify cell borders. Next, we combine those segmentation masks with the barcodes' detection confidence (which most methods produce), by simply assigning the highest-confidence barcode sequence to each cell with Some proportion of cells assigned to no barcode.

\section{Experiments}
In the following, we evaluate PLePI in the context of state-of-the-art SSOD using the MS-COCO benchmark, with CLIP~\cite{radford2021learning} as a hypothetical source of privileged information. Then using the introduced benchmarking resource in~\ref{subsec:bcbench}, we perform ablation experiments to evaluate the role of the main components in our proposed framework (PLePI-ISS) in cell calling performance. Finally, we compare recent top-performing related works on a subset of our benchmark resource.

\subsection{PLePI: MS-COCO Benchmark for SSOD}
\label{subsec:exp-coco}
Here we demonstrate PLePI on a popular SSOD benchmark as a proof of concept for incorporating independent evidence into the teacher's pseudo-labeling process. The privileged information here is provided by Contrastive Language-Image Pre-training (CLIP)~\cite{radford2021learning} model. Details on extracting evidence using CLIP for \textit{mediocre} samples and its fusion to the model's soft decisions are given in Supplementary C\textsuperscript{\ref{fn:sup}}.
We follow standard evaluation protocols for SSOD, using COCO in the most label-scarce setting. In this setting, $1\%$ of the labeled set are sampled and the remaining images are considered as unlabeled samples, which is then repeated in 5 different folds of data. We incorporated our privileged information-based enhancement of the pseudo-labeling process into PseCo~\cite{li2022pseco} which shows superior results among other methods in this minimal ratio of labeled to unlabeled samples scenario. Table~\ref{tab:coco} shows that results improve by enhancing PseCo with PLePI to make the best use of \textit{mediocre} samples according to the teacher's soft decisions (section~\ref{sec:PLePI}).

\begin{table}[ht!]
    \centering
    \begin{tabular}{lrr}
        \toprule
        Method  & $1\%$ labeled data  \\
        \midrule
        STAC~\cite{sohn2020simple}    & 13.97 $\pm$ 0.35                 \\
        Humble Teacher~\cite{tang2021humble}   & 16.96 $\pm$ 0.35         \\
        Instant-Teaching~\cite{zhou2021instant}&  18.05 $\pm$ 0.15            \\
        % Unbiased Teacher~\cite{liu2021unbiased} & 20.75 $\pm$ 0.12             \\
        Soft Teacher~\cite{xu2021end} & 20.46 $\pm$ 0.39                 \\
        Dense Teacher~\cite{zhou2022dense} & 22.38 $\pm$ 0.31                 \\
        PseCo~\cite{li2022pseco}  & 22.43 $\pm$ 0.36                      \\   
        \midrule
        PseCo + \textbf{PLePI}  & \textbf{27.70 $\pm$   0.08}          \\    
        % PseCo + \textbf{PLePI} (strong fusion)  & x               \\        
        \bottomrule
    \end{tabular}
    \caption{COCO mean average precision comparison of various state-of-the-art methods for SSOD; numbers shown as (Mean ± Standard Deviation) of the values resulted by each of 5 folds of data.}
    \label{tab:coco}
\end{table}

\subsection{PLePI-ISS: Ablation Studies for Cell Calling Evaluation}
\label{subsec:ablation}
As we are presenting the first application of SSOD to the barcode calling problem, many variations of the model's many components might be evaluated. Here we focus on the most important questions related to the contribution of this work: in the context of SSOD, is the proposed framework able to: 1) take advantage of unlabeled data and improve the initial decision boundary formed by noisy estimates? 2) take advantage of privileged information to enhance the ISS SSOD pseudo-labeling strategy?

\subsubsection{Learning/Evaluation Protocol}
We follow burn-in strategy \cite{liu2021unbiased,zhou2022dense} to initialize both the teacher and student models using labeled data.
In our data, each mini-batch contains 9 image tiles of size 256$\times$256, each containing up to 150 spots.
We evaluate the ablation experiments by randomly selecting three sets of images as labeled set, unlabeled set and test set. Labeled set consists of images in a single site of the first experimental plate, containing 4356 images of size 256$\times$256. Test set consists of images in a single site of a the second experimental plate. Unlabeled set consists of two sites of the test set plate. We evaluate the performance of applying the trained model on the test set images when using labeled data and labeled plus unlabeled data.

Using barcode calling evaluation metrics (Section~\ref{subsec:bcbench}, Supplementary A.3\textsuperscript{\ref{fn:sup}}), we report the rate of cell recovery and the goodness of the fit (as $R^2$ scores) between the image-based barcode abundance estimates and the NGS-based abundance values (Table~\ref{tab:ablation}). The table's first row is a baseline: direct application of the initialized teacher using labeled data to the test set, without taking advantage of unlabeled data.
\setlength\cmidrulewidth{0.2ex}
\begin{table}[b]
% \scriptsize
% \footnotesize
  \centering
  \resizebox{\columnwidth}{!}{%
  \begin{tabular}{l|cc|cc}
  % \toprule
    \cmidrule(r){2-5}
\multicolumn{1}{c}{} &\multicolumn{2}{c}{\footnotesize{LQ}}&\multicolumn{2}{c}{\footnotesize{HQ}}\\
% \cmidrule
\midrule
 &\footnotesize{NGS  match}    & \footnotesize{cell recovery} &  \footnotesize{NGS  match}     & \footnotesize{cell recovery} \\
% \scriptsize{(${\mathcal{L}}_{rpn},{\mathcal{L}}_{roi}$)} & \scriptsize{($\mathcal{L}_{RPN},\mathcal{L}_{ROI}$)} & 
 strategy   &  \footnotesize{($R^2$)}      &  \footnotesize{(rate)}&  \footnotesize{($R^2$)}      &  \footnotesize{(rate)} \\
\midrule

- & 0.65 & 0.54 & 0.74& 0.62   \\
    % \midrule
PLe\textcolor{gray}{PI}-ISS
    & 0.81 & 0.63  & 0.86 & 0.66   \\
PLe\textbf{PI}-ISS
    & 0.85     &   0.66 & 0.88 & 0.71 \\
    \bottomrule
  \end{tabular}
  }
  \caption{Ablation analysis of PLePI on ISS images for the barcode calling problem. Cell calling performance metrics are reported for the held out test set.}
  \label{tab:ablation}
\end{table}

\paragraph{Level of privileged information.} As emphasized in the paper, our model relies on the privileged information to avoid noise overfitting and improve its initial decision boundary. We described two types of evidence in~\ref{subsec:PIISS}, on the location and label of objects in a mini-batch of samples. We found that the application of first evidence is required to prevent failure of the pseudo-labeling process. The second row of the table corresponds to the learning strategy using the evidence for object locations only, and the third row corresponds to the proposed framework when utilizing all the evidence at hand. 

\begin{table*}[t]
% \small
  \centering
  \resizebox{\textwidth}{!}{%
  \begin{tabular}{lcc|ccc|c}
\toprule
    % \cmidrule(r){2-3} \cmidrule(r){4-5}
    
                  & NGS  match    & cell recovery
                  & $PPV$ & $FDR_{trick}$& $FDR_{other}$     & inference time
                  \\
    Method        & \scriptsize{($R^2$)}      &  \scriptsize{(rate)}
    &\scriptsize{(cell / spot)}&\scriptsize{(cell / spot)}& \scriptsize{(cell / spot)} & \scriptsize{(minutes)}
    \\
    \midrule
    % \midrule    
    ISTDECO~\cite{andersson2021istdeco}  & 0.54 & 0.45 & 0.994 / 0.992 &  0.006 /  0.0077 & 0 / 0& 75\\
    PoSTcode~\cite{gataric2021postcode}  & 0.78 & 0.71 & 0.92 / 0.93 &  0 /  0.0001 & 0.0799 /      0.0695  &  85   \\    
    Starfish~\cite{axelrod2021starfish}& 0.85 & 0.72 & 0.912 / 0.872 & 0 / 0 & 0.0883 /  0.1278&   22 \\
    BarDensr~\cite{chen2021barcode} &  0.9 & 0.71 & 0.997 / 0.99 & 0.003 /  0.0103 & 0 / 0  & 11\\
    \midrule
    \textbf{PLePI-ISS} & 0.88  & 0.72 & 0.918 /  0.908 &  0 /  0 &  0.0819 / 0.0918  & 5 \\
    \bottomrule
  \end{tabular}
  }
  \caption{Evaluation of the proposed model against SOTA baselines for data from one randomly selected site of the benchmark dataset.}
  \label{tab:baselines}
\end{table*}
% total detected cells: 8362

\paragraph{Level of annotation noise.}
As described previously, noisy annotations can be created by noisy estimations. These could come from the simplest and cheap form of thresholding (termed `low quality', LQ here) or a more costly expert-supervised parameter-tuned pipeline (`high quality', HQ). Note that both levels are considered noisy, but the goal is to evaluate the effect of using unlabeled samples and privileged information under these two annotation noise levels.

% \vspace{-.3cm}
\paragraph{Barcode Calling Results.}

Values reported in Table~\ref{tab:ablation} reveal that incorporating unlabeled samples together with any level of privileged information improves the baseline decision boundary formed by noisy labels. This benefit could be more influential under higher levels of label noise.

\subsection{Comparison to Baseline SOTA ISS Decoders}
\label{sec:bcsota}

\subsubsection{Baseline Methods}
Recent top-performing methods for decoding ISS images, as described in Section~\ref{subsec:rwiss}, find the best fit of the codebook to their real ISS data directly to estimate their proposed model's parameters. Although this strategy may cause over-detection of codebook barcodes, the lack of ground truth prevents validation of the detections in a more proper train-test evaluation scheme. The validation of their methodology usually relies on simplified simulated data with ground truth. BarDensr~\cite{chen2021barcode}, reported its performance for barcode detection in simulated data using  Receiver Operating Characteristic curve (ROC curve).  ISTDECO~\cite{andersson2021istdeco} suggests addition of extra `non-targeted' barcodes to the codebook to allow calculating a false discovery rate (FDR), defined as the ratio of non-targeted barcode detection rate to the non-target percentage of the barcodes in the codebook. Again, no unified evaluation framework exists, which hinders a fair comparison of various methodologies. 

Here, we chose four baseline methods to apply to our benchmarking ISS data. We began with the typical standard pipeline for spot-based decoding of ISS images in \textit{Starfish} package~\cite{axelrod2021starfish,ke2013situ}. As well, we test BarDensr~\cite{chen2021barcode} and ISTDECO~\cite{andersson2021istdeco} which were developed in parallel based on modeling the physical process and non-negative matrix regression methods, and the most recent work PoSTcode~\cite{gataric2021postcode}, which is based on probabilistic modeling of the data. The input to all the baseline methods is raw illumination corrected and aligned images by our in-house pipeline as described in Supplementary A\textsuperscript{\ref{fn:sup}}. 

\subsubsection{Evaluation Protocol}
As described in~\ref{subsec:ablation}, we suggest train-test schemes for evaluations to minimize the chances of false discoveries. However, not all baseline methods were developed to be performed in such a setting, so we chose one randomly selected site of the experiment as the `inference' site for all methods.
% except for our method in which we train our model on a set of labeled and unlabeled images. 
Using the available implementations for these baseline methods, we applied each method on the inference site and extracted spot-level barcode assignments. Cell-level barcode assignments were then derived by confidence measures reported for each spot-level barcode assignment as described in~\ref{sec:cellcall}.
 We note that all methods provided a barcode assignment confidence score by their defined metric except for BarDensr, for which we took the majority voting approach for spot-level to cell-level barcode assignments. As we aim to achieve the highest possible number of cells with a correct barcode assignment, the main evaluation metrics are the rate of cell recovery and the matches between the abundance of cell assignments and the NGS barcode abundance, which are reported in the first two columns of Table~\ref{tab:baselines}.
 
 Our original codebook consists of 184 barcodes (referred as \textit{targeted} barcodes) and we included 9 additional non-existent false barcodes (referred to as \textit{trick} barcodes) to the codebook for false discovery analysis. Trick barcodes are selected to have a large Hamming distance from each other and also with the targeted barcodes. Barcodes detected by each decoding method can fall into three categories: \textit{targeted}, \textit{trick} and \textit{not-targeted-nor-trick}. Targeted barcode assignment rates are reported as Positive Predictive Value (PPV) at each cell and spot level. Incorrect barcode assignment rates for two categories of \textit{trick} and \textit{not-targeted-nor-trick} calls are reported as False Discovery Rates (FDR) and are denoted as $FDR_{trick}$ and $FDR_{other}$ respectively (all metrics reported in the Table~\ref{tab:baselines} are described in details in Supplementary A.3\textsuperscript{\ref{fn:sup}}). 
 We can see that both BarDensr and ISTDECO, which force all pixels to a barcode in the codebook, have higher $FDR_{trick}$ rates. The standard ISS pipeline in Starfish has the highest overall false discovery rate, which is expected as it is the only method in the table that doesn't take advantage of the codebook in its barcode estimation process. 
 
 Although variable, we have not reported the hands-on time needed to tune parameters for each method due to the subjectivity of this task. We do report inference time for the application of each method on the test set - although our model has the minimum inference time, unlike others it requires a training phase. 

According to the numbers reported in the table, none of the methods are outperforming the rest on every single performance metric. However, our proposed method overall shows a high NGS match, while `salvaging' many cells, with a low barcode detection FDR rate, while performing in a more reliable train-test scheme.

\section{Discussion}
Alleviating the need for exhaustive human-provided labels, which may be tedious, subjective, or error-prone, is the main motivation behind semi, self, and unsupervised learning methods. These problems are explored extensively with well-defined objectives and benchmarks in the context of many common computer vision, language, and speech processing problems. However, these advancements are not always directly applicable to data from other fields of applied sciences including biology. These data can have their own unique properties, structure, and sources of noise. Furthermore, the ground truth (required for evaluation) is rarely direct, such as in the form of annotations from experts; instead, typically only indirect evidence in the form of other orthogonal data sources is available. These differences mandate a new set of assumptions and formulations. We have formulated the barcode calling problem as an object detection problem in the presence of scarce noisy labels, given a source of contextual privileged information, where a low false discovery rate is crucial. Motivated by this problem, we have proposed a general strategy for the enhancement of  pseudo-labeling scheme by a source of privileged information, which is not domain specific and can advance the general field of SSOD.

\section*{Acknowledgments}
Funding for this project was provided by Calico Life Sciences, LLC. We thank members of the labs of JT Neal, Paul Blainey, Frank Li, and Calvin Jan for their input on the project.

\bibliographystyle{named}
\bibliography{m}

\end{document}